\documentclass[letterpaper, 10 pt, conference]{ieeeconf}  % Comment this line out if you need a4paper

\IEEEoverridecommandlockouts                              % This command is only needed if 
\usepackage{amsmath} % assumes amsmath package installed
\usepackage{graphicx}
\usepackage{siunitx}
\usepackage{multirow}
\usepackage[table,xcdraw]{xcolor}
\usepackage{booktabs}

\title{\LARGE \bf
Toward Safe Autonomous Robotic Endovascular Interventions using World Models
}

\author{Harry Robertshaw$^{1}$, Nikola Fischer$^{1}$, Han-Ru Wu$^{2}$, Andrea Walker Perez$^{1}$, Weiyuan Deng$^{1}$, \\ Benjamin Jackson$^{1}$, Christos Bergeles$^{1}$, Alejandro Granados$^{1}$ and Thomas C Booth$^{1,3}$% <-this % stops a space
\thanks{$^{1}$Harry Robertshaw, Nikola Fischer, Andrea Walker Perez, Weiyuan Deng, Benjamin Jackson, Christos Bergeles, Alejandro Granados and Thomas C Booth are with Surgical \& Interventional Engineering, School of Biomedical Engineering \& Imaging Sciences, King's College London, UK}%
\thanks{$^{2}$Han-Ru WU is with the Department of  Radiology, National Taiwan University Hospital, Taiwan}%
\thanks{$^{3}$Thomas C Booth is with the Department of Neuroradiology, King's College Hospital, UK
        {\tt\small thomas.booth@kcl.ac.uk}}%
\thanks{This manuscript is a preprint and has been submitted to the IEEE/RSJ International Conference on Intelligent Robots and Systems (IROS) 2026.}
}

\begin{document}

\maketitle
\thispagestyle{empty}
\pagestyle{empty}

%%%%%%%%%%%%%%%%%%%%%%%%%%%%%%%%%%%%%%%%%%%%%%%%%%%%%%%%%%%%%%%%%%%%%%%%%%%%%%%%
\begin{abstract}

    Autonomous mechanical thrombectomy (MT) presents substantial challenges due to highly variable vascular geometries and the requirements for accurate, real-time control. While reinforcement learning (RL) has emerged as a promising paradigm for the automation of endovascular navigation, existing approaches often show limited robustness when faced with diverse patient anatomies or extended navigation horizons. In this work, we investigate a world-model-based framework for autonomous endovascular navigation built on TD-MPC2, a model-based RL method that integrates planning and learned dynamics. We evaluate a TD-MPC2 agent trained on multiple navigation tasks across hold out patient-specific vasculatures and benchmark its performance against the state-of-the-art Soft Actor-Critic (SAC) algorithm agent. Both approaches are further validated \textit{in vitro} using patient-specific vascular phantoms under fluoroscopic guidance. In simulation, TD-MPC2 demonstrates a significantly higher mean success rate than SAC ($58\%$ vs. $36\%$, $p < 0.001$), and mean tip contact forces of $0.15$\,\si{\newton}, well below the proposed $1.5$\,\si{\newton} vessel rupture threshold. Mean success rates for TD-MPC2 ($68\%$) were comparable to SAC ($60\%$) \textit{in vitro}, but TD-MPC2 achieved superior path ratios ($p = 0.017$) at the cost of longer procedure times ($p < 0.001$). Together, these results provide the first demonstration of autonomous MT navigation validated across both hold out \textit{in silico} data and fluoroscopy-guided \textit{in vitro} experiments, highlighting the promise of world models for safe and generalizable AI-assisted endovascular interventions.

\end{abstract}

%%%%%%%%%%%%%%%%%%%%%%%%%%%%%%%%%%%%%%%%%%%%%%%%%%%%%%%%%%%%%%%%%%%%%%%%%%%%%%%%
\section{INTRODUCTION}

    Stroke ranks fifth among all causes of death~\cite{Martin2024}. Mechanical thrombectomy (MT) has improved ischemic stroke outcomes, reducing mortality and disability compared to medical therapy alone~\cite{Bendszus2023}. Rapid MT intervention is necessary as effectiveness declines with delayed treatment~\cite{Asdaghi2023}. In the UK, for example, only 4.4\% of stroke admissions receive MT (far below the estimated 15\% eligibility rate) due to limited access to MT-capable centers and long inter-hospital transfers~\cite{SSNAP2025,McMeekin2024}. During MT, operator radiation exposure increases cancer and cataract risks, while protective gear contributes to orthopedic strain~\cite{Madder2017}. Robotic surgical systems could improve accessibility and reduce operator dependency~\cite{Robertshaw2023}. Tele-operated robotic MT could enable remote MT by specialists in centralized centers, while integrating AI-assistance into these systems may enable generalist interventional radiologists in peripheral hospitals (rather than specialized interventional neuroradiologists) to perform MT effectively, and could enhance efficiency and safety~\cite{Robertshaw2023}.

    Previous research into autonomous endovascular navigation has focused on aortic arch navigation~\cite{Robertshaw2023,Jianu2024,Karstensen2023}, and recent work has explored neck-to-brain navigation during MT using micro-devices~\cite{Robertshaw2025ipcai}. While these studies perform experimental evaluation over multiple vasculatures, they examine simple navigation tasks with short time episodes, which limits future clinical applicability. Larger time episodes were investigated in autonomous two-device groin-to-neck navigation during MT, but the experiments were limited to training and testing on a single vasculature~\cite{Robertshaw2024}. All these studies use reinforcement learning (RL), which can be particularly sensitive to architecture and hyperparameters and therefore are unable to perform over long time horizons~\cite{Henderson2018,Georgiev2024}. Additionally, RL is often designed for single-task learning only, limiting RL model tuning to computationally expensive models only~\cite{Hafner2023}.

    The development of policies capable of performing long navigation tasks (required for MT) across multiple patient vasculatures could be achieved with world models. These are learned representations of the environment that simulate its dynamics, enabling single agents to optimize actions in a virtual setting without relying solely on real-world data~\cite{Ha2018}. World models would allow large-scale learning from diverse datasets that could create autonomous navigation systems able to understand, predict, and adapt to real-world vascular complexities. World models with no hyperparameter tuning have been shown to outperform specialized RL agents across diverse open-source benchmark tasks~\cite{Hafner2023,Hansen2024}, although currently translation to real-world applications is limited. The model-based RL algorithm  TD-MPC2 has shown to train world models that exhibit improvements upon both the model-free Soft Actor-Critic (SAC), and the world model DreamerV3 (state-of-the-art model-based method for data-efficient continuous control~\cite{Hafner2023}) when examining multi-task environments with continuous action spaces. 
    
    A proof-of-concept study using TD-MPC2 found improvements over the state-of-the-art SAC architecture for autonomous endovascular interventions~\cite{Karstensen2025,Moosa2025} across multiple navigation tasks and vascular environments. However, there was no safety metric measurement, no generalization to hold out data, and no \textit{in vitro} validation, severely limiting clinical applicability~\cite{Robertshaw2025miccai}. The measurement of safety metrics in these studies, such as tip forces exerted by catheters and wires during navigation, is important to consider because excessive vessel wall contact forces can induce vasoconstriction and intimal damage, potentially leading to perforation or dissection-induced distal embolization and infarction in the acute setting \cite{Takashima2007}. Robotic endovascular manipulators can reduce vessel wall contact forces~\cite{RafiiTari2016}, as can the use of tip force as a loss function during autonomous navigation~\cite{Robertshaw2025ipcai}. Although the motivation for AI-assistance in MT has been to address safety concerns raised by patients regarding catheter and wire vessel wall interaction during stakeholder engagement sessions, some safety concerns persist~\cite{Robertshaw2026jaha}.
    
    While several studies have performed \textit{in vitro} testing in simplified endovascular models or idealized aortic arch anatomy~\cite{Karstensen2023,Karstensen2025,Scarponi2024}, almost all of the autonomous navigation work in the MT vasculature has taken place \textit{in silico} on patient-specific vasculatures~\cite{Robertshaw2025ipcai,Robertshaw2024,Robertshaw2025miccai}. While simulation provides a safe, scalable, and controllable environment for developing and benchmarking navigation policies, it cannot fully capture the complexities of physical interactions within vascular systems. Friction, device stiffness, and variations in catheter–guidewire mechanics can be difficult to model accurately \textit{in silico}. Physical \textit{in vitro} testing using patient-specific vasculature phantoms and fluoroscopic imaging, therefore, represents a critical step towards bridging the gap between proof-of-concept \textit{in silico} studies and clinical translation. 

    The aim of this study was to demonstrate that a singular RL agent could be used to safely perform multiple endovascular navigation tasks when moving to hold out \textit{in silico} or challenging \textit{in vitro} patient vasculatures, addressing previous shortcomings regarding generalization to unseen data and \textit{in vitro} validation. Our contributions are as follows: 1) we validated force safety metrics of the first world model capable of performing multiple endovascular navigation tasks across multiple hold out patient vasculatures \textit{in silico} for the first time; 2) we developed a clinically-relevant \textit{in vitro} testbed for MT-relevant tasks to demonstrate safe translatability, including simulation-to-real augmentation techniques; and 3) we compared our results to the current state-of-the-art RL algorithms, both \textit{in silico} and in a clinically-relevant \textit{in vitro} scenario under fluoroscopic guidance.

\section{METHODS}

    \subsection{Navigation tasks}

        In MT, a guide catheter is typically navigated from the femoral (or radial) artery to the internal carotid artery (ICA). An `access catheter' is usually placed within the guide catheter and advanced over a guidewire ahead of the guide catheter tip during navigation. Once the access catheter is within the carotid artery, the guide catheter is advanced to make a stable platform. The guidewire and access catheter are then retracted, and often a micro-guidewire within a micro-catheter can be passed through the stable guide catheter and navigated to the target thrombus site. The micro-guidewire is then removed and exchanged for a stent retriever to remove the thrombus, restoring brain blood flow.

\begin{figure}[hbt!]
            \centering
            \includegraphics[width=0.85\linewidth]{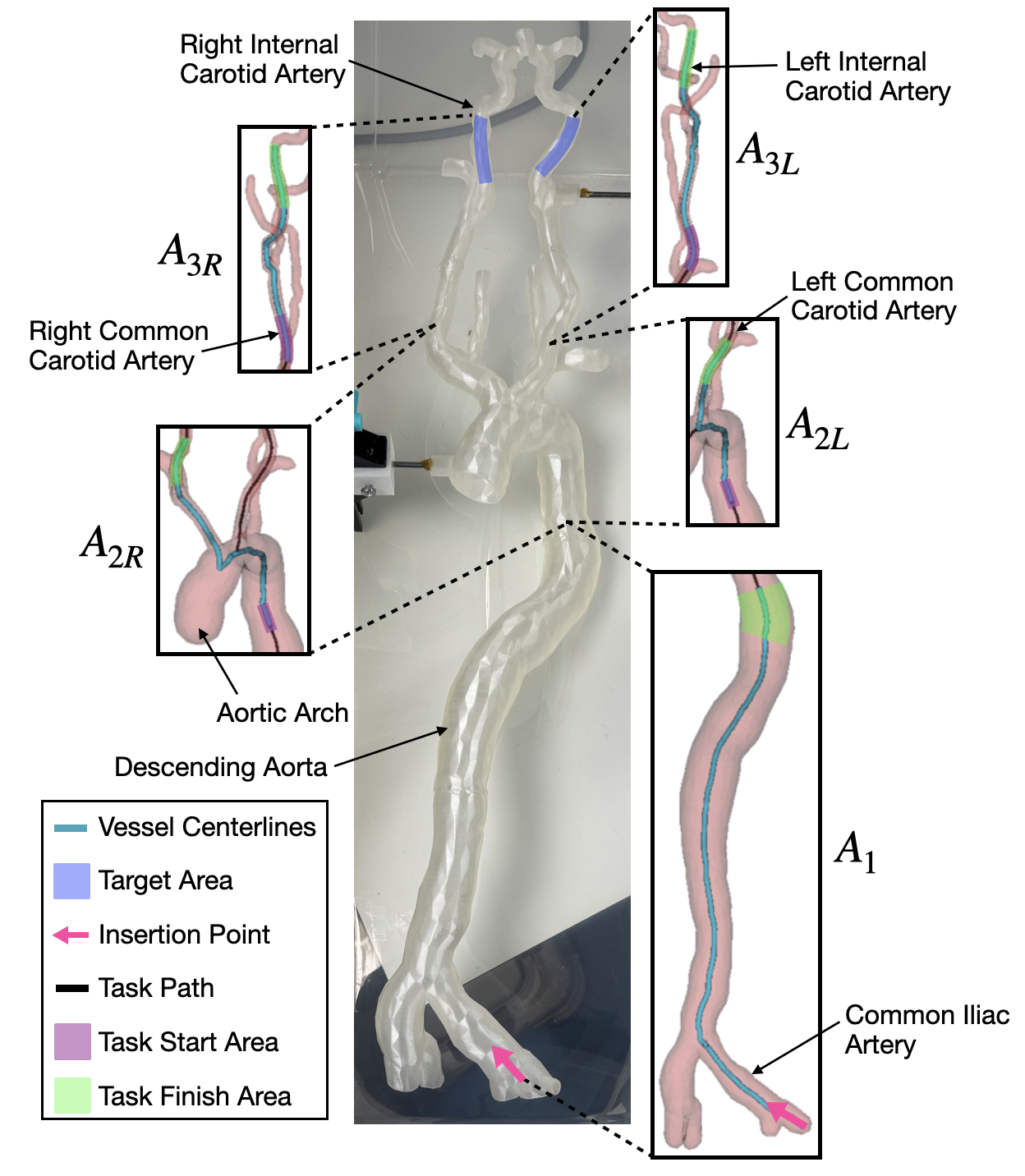}
            \caption{\textit{In vitro} 3D vascular phantom used, with first phase of thrombectomy (navigation from the femoral artery to the internal carotid artery (ICA)) navigation tasks labeled and displayed. $A_1$: Common iliac artery to superior aspect of descending aorta, $A_{2L}$: superior aspect of descending aorta to left common carotid artery (CCA), a superhuman task when using a straight catheter tip. $A_{2R}$: superior aspect of descending aorta to right CCA, $A_{3L}$: left CCA to left ICA, $A_{3R}$: right CCA to right ICA.}
            \label{fig:task_diagram}
        \end{figure}

        The first phase of MT (navigation of guidewire and guide catheter from the femoral artery to the ICA) was split into five tasks (Fig.~\ref{fig:task_diagram}) for multi-task RL training based on previous clinically motivated definitions~\cite{Robertshaw2025miccai}. A target was randomly sampled from within a range of pre-defined limits within the target vessel for each task, and the guide catheter and guidewire were navigated to it from a randomly sampled point in the starting vessel.

        The guidewire's position was described by three points equally spaced 2\,\unit{\milli\meter} apart in the distal region, denoted as $(x_g, y_g)_{i=1,2,3}$, with $(x_g, y_g)_{1}$ representing the actual tip of the guidewire~\cite{Robertshaw2024}. The final target location for navigation was specified by the coordinates $(x_t, y_t)$. Observations comprised current and previous device positions, target location, and previous action. 

    \subsection{Dataset}

        A computed tomography angiography (CTA) scan of the aortic arch and the cerebral vessels, and a CTA scan encompassing the abdominal and thoracic regions, including the femoral arteries, descending aorta, and the aortic arch (obtained with UK Research Ethics Committee 24/LO/0057), were processed into surface meshes using 3D Slicer (v5.10)~\cite{Robertshaw2025ipcai}. The centerlines and radii for all scans were extracted. The abdominal and thoracic regions were then scaled appropriately so that the radii would match at the superior aspect of the descending aorta and the inferior aspect of the aortic arch. The two sets of centerlines were joined, and the radii at multiple centerline points were used to generate a surface mesh to be loaded into the Simulation Open Framework Architecture (SOFA, v23.12)~\cite{Faure2012}. This was repeated for 15 aortic arch and cerebral vessel CTAs (15 to capture anatomy heterogeneity associated with challenging real-world navigation across diverse age/ethnicity/sex), with the same CTA of the abdominal and thoracic regions (same CTA as anatomy is homogeneous and real-world navigation is not challenging) being scaled to fit each one. Type-I aortic arches were found in 87\% (13/15) of vasculatures, while the remaining were Type-II~\cite{Lahlouh2023}. Five out of the 15 vasculatures were used as a hold out dataset for \textit{in silico} evaluation (comprising 80\% Type-I and 20\% Type-II aortic arches).

    \subsection{\textit{In silico} testbed \& data collection}

        The \textit{in silico} environment utilized the stEVE framework~\cite{Karstensen2025}. A $0.0441"$ multipurpose (MP) catheter (Terumo, Tokyo, Japan and $0.035"$ guidewire (Terumo, Tokyo, Japan) were used for all \textit{in silico} navigation tasks. The BeamAdapter plugin for SOFA was used to model these devices for the \textit{in silico} environment~\cite{Robertshaw2026ral}. This catheter type (straight) does not have a shaped tip, but it remains possible for an expert operator to navigate into the right common carotid artery (RCCA). In contrast, when this catheter type is combined with the acute angle of the left common carotid artery (LCCA), an expert operator cannot navigate the testbed – this allowed us to evaluate a superhuman additional navigation task ($A_{2L}$, Figure~\ref{fig:task_diagram}), and evaluate the full capabilities of the RL agents.
        
        The simulation assumed rigid vessel walls (with an empty lumen). The simulation's device behavior was determined by utilizing a tensile testing machine to measure the tensile strength of the devices, which facilitated the calculation of their stiffness~\cite{Jackson2023}. Friction between the wall and device was iteratively tuned to mimic behavior in a test-bench setup. This methodology has previously allowed the translation of autonomous endovascular navigation agents from \textit{in silico} to \textit{in vitro} aortic arch models~\cite{Karstensen2025}.

        Inputs (device rotation and translation speed) were applied at the proximal device end. Rotation and translation speed were constrained to a maximum of 180\,\unit{\degree\per\second} and 40\,\unit{\mm\per\second}, respectively~\cite{Karstensen2025}. Similar to a clinical scenario with fluoroscopy, feedback during the navigation was given as 2D $(x',y')$ tracking coordinates; no visual information showing the geometry of the patient vasculature was given.

    \subsection{\textit{In vitro} testbed \& robotic manipulator}

        Our \textit{in vitro} testbed consisted of a 3D-printed transparent phantom (Clear V4, Formlabs Inc., Somerville, USA), which utilized a vasculature from the \textit{in silico} training dataset (Figure~\ref{fig:task_diagram}, Type-I aortic arch). Experiments were conducted in a mock operating room at St Thomas’ Hospital, London, UK. Corresponding to clinical MT, fluoroscopic images from a mobile C-arm (Cios Spin, Siemens Healthineers, Erlangen, Germany) were used as an input to a device tip-tracking algorithm that extracted the necessary coordinates for the state-based RL algorithm~\cite{Eyberg2022}. Each task was carried out five times for both types of world model (Section~\ref{sec:RL}). Agents were allowed $5$\,\unit{\minute} during \textit{in vitro} evaluation, to account for the slower speed of the robotic manipulator (compared to \textit{in silico}).

        % typo noted after submission below

\begin{figure}[hbt!]
            \centering
            \includegraphics[width=0.7\linewidth]{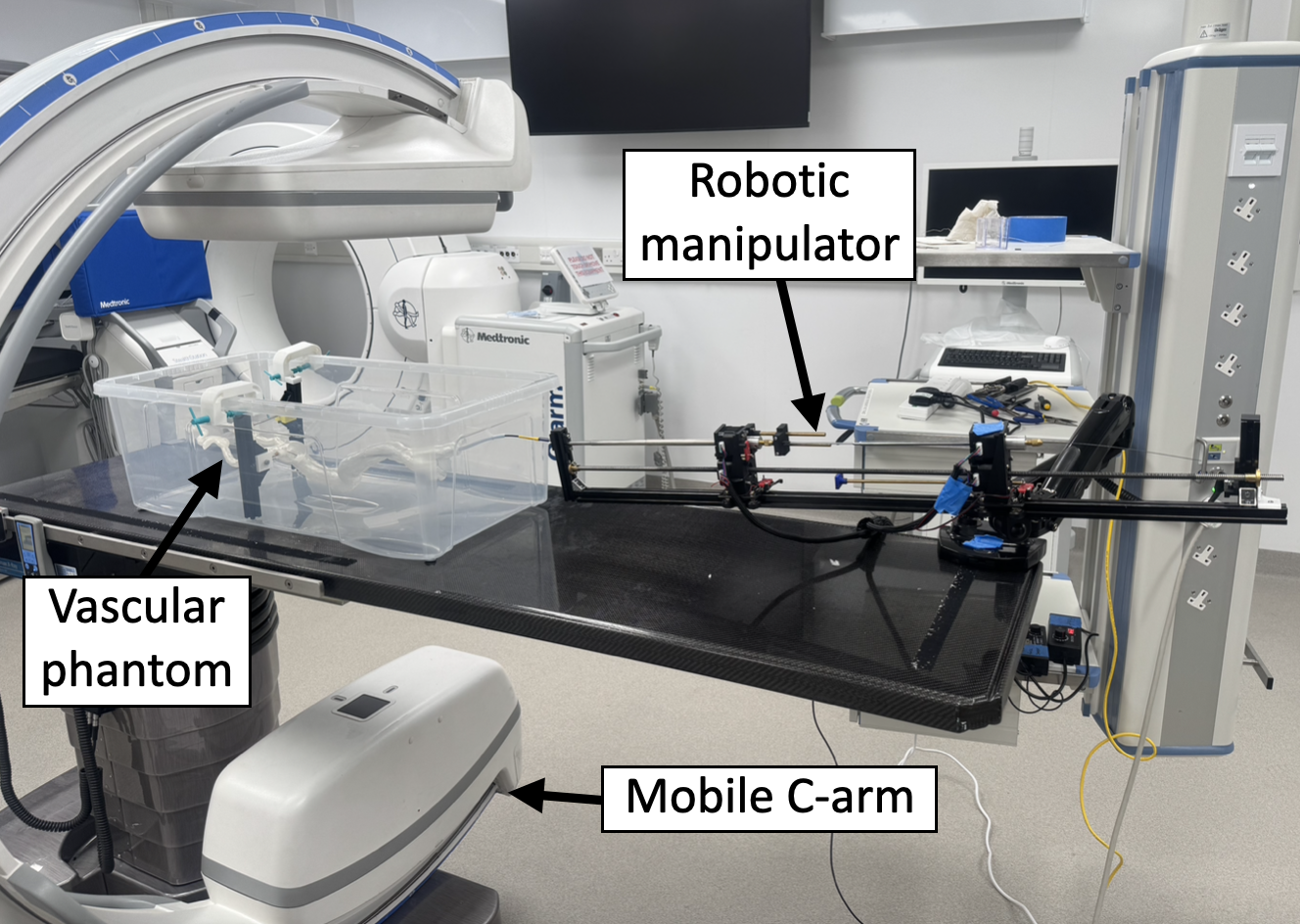}
            \caption{\textit{In vitro} testbed set-up.}
            \label{fig:in_vitro}
        \end{figure}

        A robotic manipulator (Figure~\ref{fig:in_vitro}) was used to translate and rotate the catheter and guidewire independently~\cite{Sadati2025}. The system featured independent actuation carriage units, each capable of two degrees of freedom, placed on a single lead screw rod. Direct-drive hollow shaft stepper motors were used to (1) translate the actuation carriage unit with each device (guidewire or catheter) situated along the length of the actuation system, and (2) rotate the devices via small drill chuck couplings. Telescopic tubes were installed between the carriages to prevent lateral buckling of the catheters and guidewires. As in the \textit{in silico} testbed, a $0.0441"$ multipurpose (MP) catheter (Terumo, Tokyo, Japan and $0.035"$ guidewire (Terumo, Tokyo, Japan) were used for all \textit{in vitro} navigation tasks.

    \subsection{Reinforcement learning agents}\label{sec:RL}

        Model-free SAC and model-based TD-MPC2 algorithms were used to train the RL agents in this study. The SAC RL algorithm was implemented from the open-source stEVE framework~\cite{Karstensen2025}, which has been shown to be the current state-of-the-art for autonomous endovascular interventions~\cite{Moosa2025}. In SAC, the critic learns the value and the actor optimizes the critic directly to maximize cumulative rewards~\cite{Georgiev2024}. This is useful for continuous action spaces, leveraging experience replay for data efficiency, but requires tuning and struggles under high-dimensional inputs~\cite{Hafner2023}. The SAC algorithm used here includes a Long Short-Term Memory (LSTM) layer for learning trajectory-dependent state representations and feedforward layers for controlling the devices~\cite{Karstensen2023}. 

        The TD-MPC2 algorithm is designed for sample-efficient learning and effective planning in continuous action spaces, combining temporal difference learning with model-predictive control, while leveraging a learned dynamics model to simulate environment transitions and plan action sequences over a short horizon~\cite{Hansen2024}. This approach enables the agent to optimize actions based on both predicted rewards and task constraints while using limited real-world interaction. TD-MPC2 incorporates a latent dynamics model and a cross-entropy planning method, meaning it is particularly well-suited for complex tasks with longer episode lengths. The base TD-MPC2 configuration was used but with an increased replay buffer size ($1\times10^7$), and an LSTM layer to function as an observation embedder, allowing storage of the environment state at each step so that the vessel structure could be estimated by the path of the device tip.

        The dense reward function used across training is shown in Equation~\ref{eq:R}~\cite{Robertshaw2024}. \textit{Pathlength} is defined as the distance between the guidewire tip and the target along the centerlines of the arteries, with $\Delta\text{pathlength}$ representing the change in pathlength at time $t$ from the previous step at time $t=-1$.

        \begin{equation}
            R = -0.00015 - 0.001\cdot\Delta\text{pathlength}+\begin{cases}1 & \text{if target reached} \\0 & \text{else} \end{cases}
            \label{eq:R}
        \end{equation}

    \subsection{Reinforcement learning training}

        % noticed after submission - reference shold be to miccai paper

\begin{figure}[hbt!]
            \centering
            \includegraphics[width=0.95\linewidth]{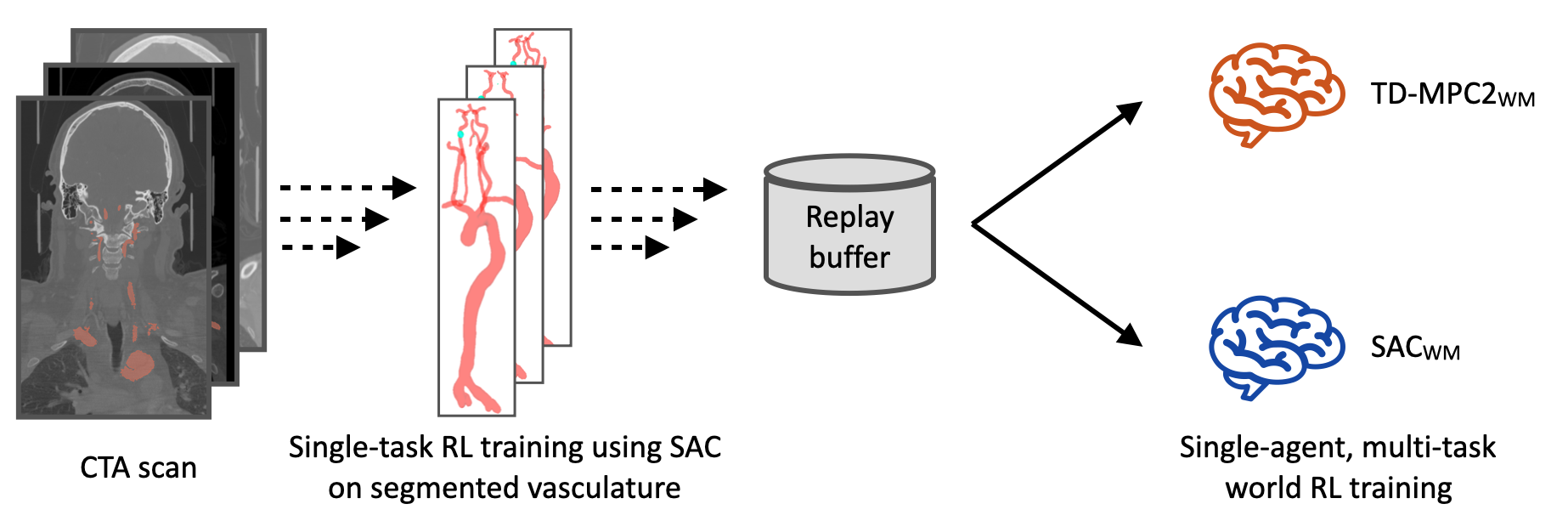}
            \caption{Training overview. An agent was trained using SAC for each task (Figure~\ref{fig:task_diagram}), and the data were used to fill the same replay buffer used for training of SAC and TD-MPC2 agents.}
            \label{fig:training}
        \end{figure}

        The RL training process is shown in Figure~\ref{fig:training}, where five agents were trained using SAC (one for each defined phase in Figure~\ref{fig:task_diagram}). Each agent performed 250 navigation episodes while recording trajectory data (trajectories, actions, rewards). This data was used to fill the replay buffers before training the multi-task agents for SAC and TD-MPC2. Each model was trained for $1 \times 10^7$ exploration steps, where a navigation task (or episode) was considered complete when the target was reached. An episode termination of 200~steps ($\approx 27$\,\unit{\second}) was set for computational efficiency. Training was performed on an NVIDIA DGX A100 node (Santa Clara, California, USA). Models were trained on 10 patient-specific vascular anatomies. All models used were trained using the same vasculatures, parameters, and SAC replay buffer data as in \cite{Robertshaw2025ipcai}, allowing for validation of this proof-of-concept study using \textit{in vitro} and hold out \textit{in silico} vasculatures, while also measuring navigation safety.

        To aid in the transition from \textit{in silico} to \textit{in vitro} testbeds, several data augmentation methods were applied during \textit{in silico} training of the SAC and TD-MPC2 agents. The entire mesh was scaled randomly from $0.7 - 1.3$ in three dimensions, and the mesh itself was rotated randomly from its origin by $\pm 30$\unit{\degree} in the $x$ and $y$ plane. Additionally, the insertion point for each task was calculated by selecting a random coordinate in the defined task start area (Figure~\ref{fig:task_diagram}), and then randomly sampling a point within the given radius at that coordinate.

    \subsection{Evaluation}

        % noticed missing word and wrong reference below after submission 
        A collision monitor on the guidewire tip was implemented to record the tip contact forces $F_{x,y,z}$ at each simulation ~\cite{Robertshaw2025ipcai}. $\|F_{x,y,z}\|$ was used to determine the maximum and the mean tip contact forces recorded during \textit{in silico} evaluation. To provide robust analysis of the previously proposed training pipeline~\cite{Robertshaw2026jaha}, the mean/maximum guidewire tip speed was also recorded during \textit{in silico} evaluation. Additionally, an ablation study on the effectiveness of the applied data augmentation techniques was carried out to determine their effectiveness in translating agents from \textit{in silico} to \textit{in vitro}. The ablation study was done \textit{in vitro} using the multi-task SAC agent on $A_{2R}$.

        During both \textit{in silico} and \textit{in vitro} evaluation the success rate (percentage of evaluation episodes where the target is reached), procedure time (time from the start of navigation to target point for successful episodes), path ratio (percentage distance navigated to target location in unsuccessful episodes, calculated by dividing guidewire tip total distance traveled towards target point by the initial distance from guidewire tip to target point) and exploration steps (number of training steps taken to reach the point at which the results are provided) were recorded. Across all experiments and metrics, reported values were mean $\pm$ standard deviation, while comparative statistical analyses were conducted using two-tailed paired Student’s $t$-tests with a predetermined significance threshold set at $p = 0.05$.
    
\section{RESULTS}

    \subsection{\textit{In silico} evaluation}

        Results from \textit{in silico} evaluation (Table~\ref{tab:world_task_merged}) favored TD-MPC2 over SAC with an increase in mean success rate (58\% to 36\% [$p< 0.001$]) and mean path ratio (22\% to 49\% [$p< 0.001$]), although taking nearly twice as long (17.1\,s to 9.6\,s [$p< 0.001$]). Increases in task success rate and path ratio were seen when moving from a SAC to TD-MPC2 agent: $A_{2L}$ (SR: $8\%$ to $38\%$ [$p< 0.001$], PR: $21\%$ to $40\%$ [$p< 0.001$]), $A_{2R}$ (SR: $8\%$ to $52\%$ [$p< 0.001$], PR: $24\%$ to $48\%$ [$p< 0.001$]), and $A_{3L}$ (SR: $16\%$ to $38\%$ [$p= 0.015$], PR: $21\%$ to $63\%$ [$p< 0.001$]). An increase in procedure time was observed across $A_{1}$ and $A_{3R}$ when moving from SAC to TD-MPC2. 

        An increase in mean and maximum tip force was found for the TD-MPC2 navigations (mean: $p< 0.040$, max: $p< 0.005$). No statistically significant difference was found between the examined models for mean or maximum tip speed. Exploration steps recorded for SAC and TD-MPC2 were $4.5\times10^6$ (reached after 75~hours) and $0.5\times10^6$ (reached after 25~hours), respectively. 

        \begin{table*}[hbt!]
            \centering
            \footnotesize
            \caption{In silico world model performance and guidewire tip safety metrics on the hold-out set. Arrows (↑/↓) indicate whether higher or lower values are preferred, respectively. Tip force and speed are reported as mean and maximum values, averaged over all evaluation episodes for a given task. Asterisks denote significant differences between TD-MPC2 and SAC (* $p < 0.05$, ** $p < 0.01$, *** $p < 0.001$). Mean row (bottom) shows averages across all tasks ($n = 250$).}
            \label{tab:world_task_merged}
            \begin{tabular}{c c c c c c c c c}
                \toprule
                \textbf{Task} 
                & \textbf{Model} 
                & \textbf{\begin{tabular}[c]{@{}c@{}}Success \\ Rate \\ (\%, ↑)\end{tabular}}
                & \textbf{\begin{tabular}[c]{@{}c@{}}Procedure \\ Time \\ (s, ↓)\end{tabular}}
                & \textbf{\begin{tabular}[c]{@{}c@{}}Path \\ Ratio \\ (\%, ↑)\end{tabular}}
                & \textbf{\begin{tabular}[c]{@{}c@{}}Tip Force \\ Mean (N, ↓)\end{tabular}}
                & \textbf{\begin{tabular}[c]{@{}c@{}}Tip Force \\ Max (N, ↓)\end{tabular}}
                & \textbf{\begin{tabular}[c]{@{}c@{}}Tip Speed \\ Mean (mm/s, ↓)\end{tabular}}
                & \textbf{\begin{tabular}[c]{@{}c@{}}Tip Speed \\ Max (mm/s, ↓)\end{tabular}} \\
                \midrule
                $A_1$   
                & SAC     
                & 94 $\pm$ 24
                & 8.1 $\pm$ 0.9
                & $25 \pm 9$ 
                & 0.10 $\pm$ 0.04
                & 0.35 $\pm$ 0.05
                & 41.3 $\pm$ 3.0
                & 85.0 $\pm$ 21.8 \\
                & TD-MPC2
                & 94 $\pm$ 24
                & 15.4 $\pm$ 2.8*** 
                & 31 $\pm$ 18
                & 0.12 $\pm$ 0.04
                & 0.40 $\pm$ 0.10***
                & 24.3 $\pm$ 3.2***
                & 54.4 $\pm$ 11.6*** \\
                \midrule
                $A_{2L}$ 
                & SAC     
                & $8 \pm 27$ 
                & 15.4 $\pm$ 11.3
                & $21 \pm 11$ 
                & 0.12 $\pm$ 0.12
                & 0.66 $\pm$ 0.29
                & 14.6 $\pm$ 7.6
                & 90.2 $\pm$ 44.5 \\
                & TD-MPC2
                & 38 $\pm$ 49*** 
                & 17.8 $\pm$ 7.5 
                & 40 $\pm$ 18*** 
                & 0.13 $\pm$ 0.04
                & 0.60 $\pm$ 0.20
                & 27.3 $\pm$ 6.1***
                & 95.9 $\pm$ 44.7 \\
                \midrule
                $A_{2R}$ 
                & SAC     
                & $8 \pm 27$ 
                & 12.3 $\pm$ 5.8
                & $24 \pm 14$ 
                & 0.13 $\pm$ 0.06
                & 0.54 $\pm$ 0.21
                & 17.9 $\pm$ 12.3
                & 88.0 $\pm$ 80.8 \\
                & TD-MPC2
                & 52 $\pm$ 50*** 
                & 20.2 $\pm$ 6.8
                & 48 $\pm$ 21*** 
                & 0.14 $\pm$ 0.05
                & 0.63 $\pm$ 0.18*
                & 25.6 $\pm$ 2.9***
                & 95.0 $\pm$ 45.1 \\
                \midrule
                $A_{3L}$ 
                & SAC    
                & $16 \pm 37$ 
                & 25.2 $\pm$ 7.9 
                & $21 \pm 21$ 
                & 0.14 $\pm$ 0.08
                & 0.50 $\pm$ 0.18
                & 17.3 $\pm$ 5.4
                & 56.9 $\pm$ 36.8 \\
                & TD-MPC2 
                & 38 $\pm$ 49* 
                & 21.1 $\pm$ 6.7
                & 63 $\pm$ 18*** 
                & 0.16 $\pm$ 0.07
                & 0.54 $\pm$ 0.12
                & 23.3 $\pm$ 5.4***
                & 77.0 $\pm$ 62.0 \\
                \midrule
                $A_{3R}$ 
                & SAC    
                & $54 \pm 50$ 
                & 6.4 $\pm$ 5.7
                & $20 \pm 30$ 
                & 0.18 $\pm$ 0.13
                & 0.46 $\pm$ 0.14
                & 34.7 $\pm$ 23.7
                & 93.9 $\pm$ 112.0 \\
                & TD-MPC2
                & 66 $\pm$ 48
                & 14.6 $\pm$ 6.9*** 
                & 44 $\pm$ 21* 
                & 0.19 $\pm$ 0.05
                & 0.57 $\pm$ 0.10**
                & 24.8 $\pm$ 8.8***
                & 71.1 $\pm$ 54.9 \\
                \midrule
                \rowcolor{gray!10}
                \textbf{Mean} 
                & SAC   
                & $36 \pm 48$ 
                & 9.6 $\pm$ 6.9
                & $22 \pm 19$ 
                & 0.13 $\pm$ 0.09
                & 0.50 $\pm$ 0.22
                & 24.6 $\pm$ 16.0
                & 82.2 $\pm$ 65.9 \\
                \rowcolor{gray!10}
                & TD-MPC2
                & 58 $\pm$ 50*** 
                & 17.1 $\pm$ 6.4*** 
                & 49 $\pm$ 21*** 
                & 0.15 $\pm$ 0.06*
                & 0.55 $\pm$ 0.16**
                & 25.1 $\pm$ 5.8
                & 78.9 $\pm$ 50.0 \\
                \bottomrule
            \end{tabular}
        \end{table*}

    \subsection{\textit{In vitro} evaluation}

        % need to reference table in text

        Evaluating the trained models on an \textit{in vitro} testbed showed no difference in mean success rate between TD-MPC2 and SAC ($p= 0.567$). Additionally, no significant difference was recorded across any of the individual tasks despite higher success rates for TD-MPC2 compared to SAC in $A_{2R}$ ($60\%$ to $40\%$ [$p= 0.581$]) and $A_{3R}$ ($80\%$ to $60\%$ [$p= 0.547$]). An increase in path ratio was observed when moving from SAC to TD-MPC2 for $A_{2L}$ ($47\%$ to $71\%$ [$p= 0.017$]). Mean procedure time was significantly quicker in SAC across tasks (TD-MPC2: 111.0\,\si{\second} to SAC: 66.2\,\si{\second} [$p< 0.001$]). A significant increase in procedure time was observed when moving from \textit{in silico} to \textit{in vitro} for both TD-MPC2 (mean time \textit{in silico}: $17.1$\,\unit{\second}, \textit{in vitro}: $111.0$\,\unit{\second} [$p<0.001$]) and SAC (mean time \textit{in silico}: $9.6$\,\unit{\second}, \textit{in vitro}: $66.2$\,\unit{\second} [$p<0.001$]). This increase is caused by the delay between each step (0.5\,\unit{\second}) necessary for the robot to complete its movement, and to ensure that the motors did not overheat.

        \begin{table}[hbt!]
            \centering
            \scriptsize
            \caption{In vitro world model performance. Arrows (↑/↓) indicate whether higher or lower values are preferred, respectively. Asterisks denote significant differences between TD-MPC2 and SAC (* $p < 0.05$, ** $p < 0.01$, *** $p < 0.001$). Mean row (bottom) shows averages across all tasks ($n = 25$).}
            \label{tab:world_task_vitro}
            \begin{tabular}{c c c c c}
                \toprule
                \textbf{Task}
                & \textbf{Model}
                & \textbf{\begin{tabular}[c]{@{}c@{}}Success \\ Rate (\%, ↑)\end{tabular}}
                & \textbf{\begin{tabular}[c]{@{}c@{}}Procedure \\ Time (s, ↓)\end{tabular}}
                & \textbf{\begin{tabular}[c]{@{}c@{}}Path \\ Ratio (\%, ↑)\end{tabular}} \\
                \midrule
                $A_1$   
                & SAC     & 100 $\pm$ 0 & 46.2 $\pm$ 3.6 & -- \\
                & TD-MPC2 & 100 $\pm$ 0 & $80.4 \pm 19.7^{**}$ & -- \\
                \midrule
                $A_{2L}$ 
                & SAC     & 0 $\pm$ 0 & -- & $47 \pm 11$ \\
                & TD-MPC2 & 0 $\pm$ 0 & -- & 71 $\pm$ 14$^{*}$ \\
                \midrule
                $A_{2R}$ 
                & SAC     & $40 \pm 55$ & 97.0 $\pm$ 1.4 & $42 \pm 4$ \\
                & TD-MPC2 & 60 $\pm$ 55 & $125.0 \pm 21.0$ & 49 $\pm$ 7 \\
                \midrule
                $A_{3L}$ 
                & SAC     & 100 $\pm$ 0 & 54.4 $\pm$ 1.52 & -- \\
                & TD-MPC2 & 100 $\pm$ 1.5 & $126.4 \pm 57.7^{*}$ & -- \\
                \midrule
                $A_{3R}$ 
                & SAC     & 60 $\pm$ 55 & 67.0 $\pm$ 7.2 & 45 $\pm$ 30 \\
                & TD-MPC2 & 80 $\pm$ 45 & $112.3 \pm 7.5^{***}$ & $43 \pm 0$ \\
                \midrule
                \rowcolor{gray!10}
                \textbf{Mean}
                & SAC     & $60 \pm 50$ & 66.2 $\pm$ 17.2 & $44 \pm 13$ \\
                \rowcolor{gray!10}
                & TD-MPC2 & 68 $\pm$ 48 & $111.0 \pm 37.3^{***}$ & 54 $\pm$ 17$^{*}$ \\
                \bottomrule
            \end{tabular}
        \end{table}

\begin{figure}[hbt!]
            \centering
            \includegraphics[width=0.8\linewidth]{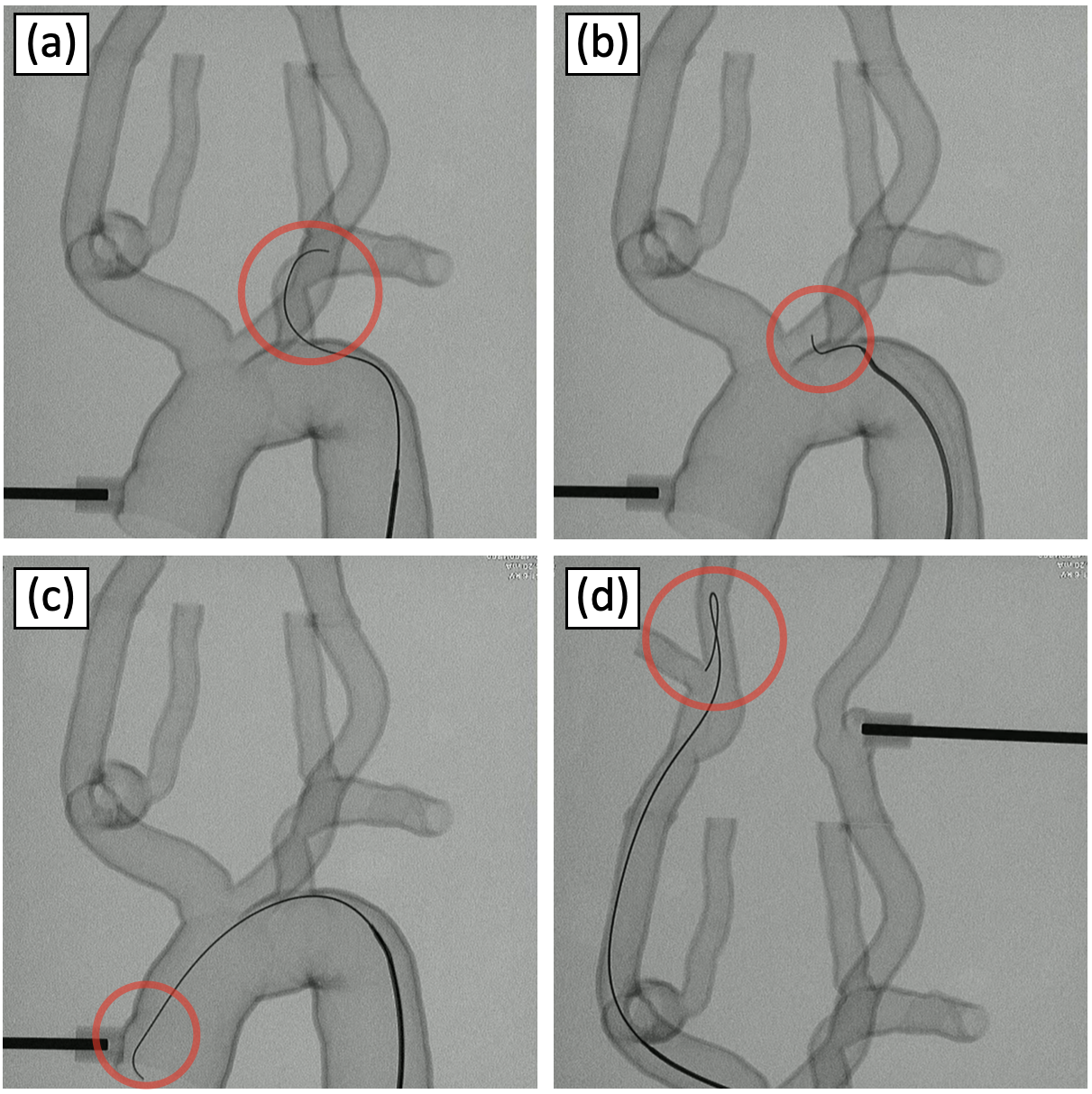}
            \caption{Examples of \textit{in vitro} failure modes with areas highlighted by red circle. (a) Wrong branch catheterization during $A_{2L}$ - the wire has entered the left subclavian artery, (b) entering the LCCA during $A_{2L}$ but not utilizing the catheter effectively to reach the target further up the branch, (c) device leaving the phantom through the origin of the ascending aorta during $A_{2R}$, and (d) unable to recover from device coiling during $A_{3R}$.}
            \label{fig:failure_modes}
        \end{figure}

        Failure modes using SAC were due to: wrong branch catheterization ($40\%$, $4/10$), episode timeout at $5$\,mins ($30\%$, $3/10$), device leaving the phantom at the origin of the ascending aorta ($20\%$, $2/10$), and unable to recover from device coiling ($10\%$, $1/10$). The TD-MPC2 failure modes were: episode timeout ($50\%$, $4/8$), entering the LCCA during $A_{2L}$ but not utilizing the catheter effectively to reach the target further up the branch ($25\%$, $2/8$), wrong branch catheterization ($13\%$, $1/9$), and device leaving the phantom at the origin of the ascending aorta ($13\%$, $1/10$). Example failure modes can be seen in Figure~\ref{fig:failure_modes}.

        The \textit{in vitro} ablation experiment of data augmentation (Table~\ref{tab:aug_vitro}) showed that augmentation provided non-significant improvement ($40\%$ to $20\%$ [$p= 0.549$]). Failure modes during navigation by the agent with no augmentation applied were all due to timeout, as it was not able to catheterize the correct branch and remained in the aortic arch.

        \begin{table}[hbt!]
            \centering
            \scriptsize
            \caption{Model comparison with and without augmentation applied for task $A_{2R}$ \textit{in vitro}. Arrows (↑/↓) indicate whether higher or lower values are preferred, respectively.}
            \label{tab:aug_vitro}
            \begin{tabular}{c c c c}
                \toprule
                \textbf{Model}
                & \textbf{\begin{tabular}[c]{@{}c@{}}Success \\ Rate (\%, ↑)\end{tabular}}
                & \textbf{\begin{tabular}[c]{@{}c@{}}Procedure \\ Time (s, ↓)\end{tabular}}
                & \textbf{\begin{tabular}[c]{@{}c@{}}Path \\ Ratio (\%, ↑)\end{tabular}} \\
                \midrule
                SAC no augmentation
                & $20 \pm 55$ & 88.0 $\pm$ 0 & $41 \pm 14$ \\
                \midrule
                SAC with augmentation
                & 40 $\pm$ 55 & 97.0 $\pm$ 1.4 & 42 $\pm$ 4 \\
                \bottomrule
            \end{tabular}
        \end{table}

\section{DISCUSSION}

    % typo below noticed after submission
    This study demonstrates notable progress in the development of RL models for autonomous MT, addressing multiple navigation tasks with industry-standard devices tracked using fluoroscopy across five hold out real patient vasculatures \textit{in silico} and a
    n\textit{in vitro} testbed, for the first time. The difference in performance between $A_{2L}$ and $A_{2R}$ highlights potential challenges related to increased vasculature complexity. This variation requires further exploration to identify anatomical features that influence performance, but reflects clinical scenarios where the left CCA is typically harder to catheterize than the right, and may need specialist access catheters with angled tips, which we describe below.

    The results demonstrate a clear advantage of TD-MPC2 over SAC in hold out \textit{in silico} data, with an increase in mean success rate observed when using TD-MPC2. For individual tasks, TD-MPC2 significantly outperformed SAC \textit{in silico} across $A_{2L}$, $A_{2R}$, and $A_{3L}$, while recording the same or non-significant increases in success rate for $A_{1}$ and $A_{3R}$. This highlights the advantages of TD-MPC2, which leverages a learned dynamics model and horizon-based planning to optimize actions in diverse and complex environments. SAC’s model-free approach may struggle to generalize effectively to anatomical variations and multiple objectives. While both SAC and TD-MPC2 failed to solve all tasks they were trained on, the findings suggest that TD-MPC2 is better suited for scenarios requiring multi-task generalization and adaptability to unseen environments. 
    
    Although gains in success rate and path ratio were recorded, TD-MPC2 resulted in increased procedure time across all tasks. This trade-off suggests that TD-MPC2 may prioritize more cautious exploration and deliberate navigation strategies, improving overall success but at the expense of efficiency. Whilst future work could explore optimizing the trade-off between efficiency and exploration in navigation to be similar to SAC, the time difference (seconds) is negligible compared to the time for patient transfer (hours), which this technology aims to obviate. Additionally, while significant increases in mean and maximum force were recorded across tasks for TD-MPC2, all values recorded are well below a proposed vessel rupture threshold of 1.5\,\si{\newton}~\cite{Jackson2023}. As the mean tip force for TD-MPC2 across all tasks is $10\%$ of the threshold value, the small increase in force is acceptable given the large increase in success rate it is associated with.

    % typo below

    When transitioning to \textit{in vitro} evaluation, a similar pattern of performance was observed between the two algorithms. Procedure times remained longer, and path ratios decreased under TD-MPC2 across tasks. Although there appeared to be small increases in success rate with TD-MPC2, the results were non-significant, which may reflect the limited number of evaluations performed per navigation task ($n=5$). In some cases, differences success rates were recorded across \textit{in vitro} tasks compared to \textit{in silico}, potentially due to manufacturing differences that arose when transferring the simulation-based vasculature to a physical phantom.

    Other work has investigated world models for this task, but used the same data for training and testing~\cite{Robertshaw2025miccai}. Here, we present comparable mean success rates for TD-MPC2 across all tasks but on hold out data ($65\%$~\cite{Robertshaw2025miccai} vs $58\%$ (ours) [$p=0.111$]), suggesting that the proposed approach can also generalize navigation to unseen data with minimal performance decrease. Although the \textit{in silico} success rate of $58\%$ for TD-MPC2 across hold out patient vasculatures suggests the need for further optimization, it represents a step toward bridging the gap between experimental performance and clinical feasibility. SAC demonstrated slightly less applicability to more real-world scenarios, where agents must navigate diverse vasculatures.

    Both of the trained models were unable to perform $A_{2L}$ \textit{in vitro}. Increasing vessel tortuosity and unfavorable carotid artery take-off angles from the aortic arch can increase endovascular navigation difficulty~\cite{Shazeeb2023}. Techniques used to navigate the LCCA \textit{in silico} were not able to be replicated \textit{in vitro}, potentially due to differences in friction and device properties. An expert interventional neuroradiologist (10 years as UK consultant; US attending equivalent) had experimented on the testbed and confirmed that in a clinical scenario, specialist catheters would be used, namely a guide catheter with a slightly bent tip, or a shaped slip catheter such as a Simmons 2, which has a reverse curve. In summary, in the current work we have also shown that a single \textit{in silico} trained world model can be effectively transferred to an \textit{in vitro} testbed to complete multiple, non-superhuman (i.e. not $A_{2L}$), MT-relevant navigation tasks autonomously. 

    Several limitations should be acknowledged. The current evaluation was performed using one \textit{in vitro} model, which may limit the generalizability of the findings to broader anatomical variations and use cases beyond what we describe here. The training dataset of ten cases could also be expanded to include a larger representation of varying vascular complexities to increase generalization.

    Future work should focus on testing the proposed approach on a range of hold out \textit{in vitro} vasculatures, while looking for performance improvements on prior, benchmarked datasets. Other world models have demonstrated scalability to 80-150 non-medical benchmark tasks~\cite{Georgiev2024,Hafner2023,Hansen2024}, demonstrating potential for future navigation applications as increasingly diverse vasculatures are incorporated within datasets. Future work should also look to measure forces exhibited \textit{in vitro} vessel walls by, for example, using motor current as a proxy for axial or rotational load, or possibly through analyzing the physician’s muscular activity and motion patterns~\cite{Sierotowicz2025}.

\section{CONCLUSIONS}

    This study demonstrates the feasibility of a world model-based RL approach for autonomous MT, addressing key challenges in generalization and multi-task learning. The proposed world model leverages existing open-source repositories to maximize reproducibility~\cite{Karstensen2025,Hansen2024}. TD-MPC2 achieved superior generalization across hold out data compared to SAC, with an increase in success rate and path ratio \textit{in silico}. This is the first validation of an autonomous MT navigation agent in an \textit{in vitro} testbed under fluoroscopic guidance. While improvements in generalization are needed before clinical deployment, the \textit{in vitro} results confirm that agents trained \textit{in silico} can navigate multiple tasks in physical models. The findings presented here provide a critical step toward the development of AI-driven autonomous robotic endovascular interventions with improved safety and precision.

\addtolength{\textheight}{-5cm}   % This command serves to balance the column lengths

%%%%%%%%%%%%%%%%%%%%%%%%%%%%%%%%%%%%%%%%%%%%%%%%%%%%%%%%%%%%%%%%%%%%%%%%%%%%%%%%

%%%%%%%%%%%%%%%%%%%%%%%%%%%%%%%%%%%%%%%%%%%%%%%%%%%%%%%%%%%%%%%%%%%%%%%%%%%%%%%%



\section*{ACKNOWLEDGMENT}

Partial financial support was received from the WELLCOME TRUST (203148/A/16/Z), the EPSRC Doctoral Training Partnership (EP/R513064/1), and the MRC IAA 2021 King's College London (MR/X502923/1). The authors thank the staff and facilities at Surgical \& Interventional Engineering (SIE), KCL, for their support. Work in the SIE Validation Suite has been supported by the Wellcome/EPSRC Centre for Medical Engineering [WT203148/Z/16/Z], a multi-user equipment grant for post-mortem evaluation of medical devices from Wellcome [218286/Z/19/Z], and the Wolfson Foundation [PR/ylr/md/21896]. For the purpose of Open Access, the Author has applied a CC BY public copyright license to any Author Accepted Manuscript version arising from this submission.

%%%%%%%%%%%%%%%%%%%%%%%%%%%%%%%%%%%%%%%%%%%%%%%%%%%%%%%%%%%%%%%%%%%%%%%%%%%%%%%%

\bibliographystyle{IEEEtran}
\bibliography{IEEEabrv,mybibfile}

\end{document}